# Critical biblical studies via word frequency analysis: unveiling text authorship

Shira Faigenbaum-Golovin[1,¶,*], Alon Kipnis[2,¶], Axel Bühler[3], Eli Piasetzky[4,&], Thomas Römer[5,&] and Israel Finkelstein[6,&]

October 23, 2024

**Affiliations**

[1] Department of Mathematics and Rhodes Information Initiative, Duke University, NC, USA.
[2] School of Computer Science, Reichman University, Herzliya, Israel.
[3] Department of Biblical Texts, Protestant Faculty of Theology of Paris, France.
[4] School of Physics and Astronomy, Sackler Faculty of Exact Sciences, Tel Aviv University, Israel.
[5] Chair of Northwest Semitic and Biblical Studies, Collège de France, France.
[6] School of Archaeology and Maritime Cultures, University of Haifa, Israel.
[¶] Authors contributed equally to this work.
[&] Authors co-led the research team.

## Abstract

The Bible, a product of an extensive and intricate process of oral-written transmission spanning centuries, obscures the contours of its earlier recensions. Debate rages over determining the existing layers and identifying the date of composition and historical background of the biblical texts. Traditional manual methodologies have grappled with authorship challenges through scrupulous textual criticism, employing linguistic, stylistic, inner-biblical, and historical criteria. Despite recent progress in computer-assisted analysis, many patterns still need to be uncovered in Biblical Texts. In this study, we address the question of authorship of biblical texts by employing statistical analysis to the frequency of words using a method that is particularly sensitive to deviations in frequencies associated with a few words out of potentially many. In this study, the term "word" may be generalized to "word n-gram" or other countable text features. We aim to differentiate between three distinct "authors" across numerous chapters spanning the first nine books of the Bible. In particular, we examine 50 chapters labeled according to biblical exegesis considerations into three corpora: the old layer in Deuteronomy, texts belonging to the "Deuteronomistic History" in Joshua-to-Kings, and Priestly writings. Without prior assumptions about author identity, our approach leverages subtle differences in word frequencies to distinguish among the three corpora and identify author-dependent linguistic properties. Our analysis indicates that the first two authors (D and DtrH) are much more closely related compared to P, a fact that aligns with expert assessments. Additionally, we attain high accuracy in attributing authorship by evaluating the similarity of each chapter with the reference corpora. In a secondary, using the authorship of the three corpora as ground truth, we investigate other biblical texts of disputed authorship among biblical experts. This study sheds new light on the authorship of biblical texts by providing interpretable, statistically significant evidence that there are different linguistic characteristics of biblical authors and that these differences can be identified. Our methodology thus provides a new tool to address contentious matters in biblical studies.

# 1. Introduction

The Hebrew Bible is a product of a centuries-long process, of texts written by different authors [1-5]. One of the main questions raised in biblical studies is identifying the authorship of biblical texts. Such an identification can shed light on historical understanding, literary analysis, textual criticism, and theological interpretation. Therefore, identifying the authorship of biblical texts is crucial for enriching our understanding of the Bible as a historical document, a work of literature with profound theological significance, and helps scholars assess the reliability and authenticity of the text. Traditionally, scholars identify the authorship of biblical texts based on language, style, inner-biblical logic, and geographical and historical information. Such approaches have managed to distinguish among several main biblical authors and identify the authorship of most pieces. Nevertheless, despite thousands of years of studies, additional authorship challenges concerning the bible remain.

With the rise of Digital Humanities, numerous methods were developed to tackle the question of authorship in modern and ancient texts [6-17] (see additional discussion in the Supplementary Material). A general authorship challenge involves data consisting of several homogenous corpora of known authorship. Upon introducing a new document of unknown authorship, we wish to associate it with one of the known-authorship corpora or conclude that the author of the new document is unlikely to be represented in the data [18]. Since a single author can write multiple texts on different topics using different genres [19], a major challenge in authorship studies is the identification of features characteristics of the author but relatively unrelated to the topic or genre [6,11]. Several recent studies applied modern techniques in machine learning and Natural Language Processing (NLP), such as Artificial Neural Networks (NN) [20-21]. However, despite apparent progress in authorship attribution methods and their application to the bible [9-10, 14-17], there are unresolved challenges.

Distinguishing among authors and attributing authorship in the bible is particularly challenging for several reasons. First, most of the texts are multilayered (edited, and re-edited) over a long time period, and hence pose a challenge for characterizing the primary composition. Additionally, very few ground-truth data are agreed upon among scholars, therefore supervised learning techniques are typically unsuitable as they require substantial amounts of ground-truth data. Analyzing biblical texts in their original Hebrew version is preferable for information perspective, but presents challenges to practitioners. Finally, as the bible has been studied for thousands of years, insights provided by an authorship study is arguably only useful to bible scholars when the involved tools are transparent and explainable. The chapter system used today, originated in medieval times and therefore cannot be taken as a basis for authorship consideration. Moreover, the chapters are very short and pose a problem for robust statistical inference. In addition, there exist challenges that are solely introduced by the Hebrew language; relying on the English translation of biblical texts can obscure author-characteristic nuances. Thus, existing methods are challenged by the properties of Biblical texts.

In this study, we harness a distinctive collaboration between biblical experts and statisticians to address the following questions quantitatively: we wish to determine whether distinct linguistic properties exist that characterize authors in the Bible. In addition, to ascertain whether these properties can be identified and utilized to estimate the likelihood of attributing a new chapter to one of the known authors.

The key contribution of this paper is a new analysis of authorship in the Bible based on lemma n-gram frequencies via a method that is sensitive to differences between authors associated with potentially rare and subtle deviations of their occurrences. This method, introduced in [22, 23], has interesting optimality properties [24] and is based on the notion of Higher Criticism in statistics [25-26]. Unlike previous studies relying on word n-gram frequencies [6,11,27], this method has no tunable parameters and does not require reducing to a small set of handcrafted features. This ensures that the results are interpretable and free from the researcher's subjective influence. Furthermore, the method also identifies a set of lemma n-gram providing the best evidence for authorship discrimination, thus providing a valuable interpretation to authorship similarity or discrimination.

We propose an algorithmic framework that adapts this method to existing authorship challenges in the bible. Our analysis results not only with attribution likelihood but also a list of discriminating lemmas provided by our framework in the context of biblical scholarship. An additional contribution of this paper is the sensitivity analysis of the method, which had not been previously performed. For example, it was unclear how well the method could be applied to short texts, whether it depends on the reference corpora, and how sensitive it is to the statistical properties of words. In this paper, we test the method under these scenarios and characterize its success rate in Section 2.4 below.

The study outlined in this paper consists of two parts. In the first part, we focus on the first nine books of the Bible. Based on conventional wisdom in biblical exegesis, we assembled three groups of chapters (referred to below as "corpora") to serve as ground truth data for this study: (a) *The old layer in Deuteronomy* (hereafter D), usually dated to late monarchic times in the late 7th century BCE (e.g., [28] p. 231; [29] p. 34); (b) *The early layer in the Deuteronomistic History* (DtrH – this term applies to the Books of Joshua, Judges, 1 and 2 Samuel, and 1 and 2 Kings), dated to late monarchic times in the late 7th century or to the exilic period in the 6th century BCE [4]; (c) *Priestly writings* (P), which are commonly dated to late exilic or post-exilic times (see [30], p. 93). We assume that each corpus represents an author or authors of the same cultural/ideological background. Although one cannot assume these chapters to be free of editorial layers (for the DtrH see [4]), there is a general agreement that an important revision of Deuteronomy took place when the book became part of the Dtr History or Dtr Library. However, there is also consensus that the so-called DtrH is multilayered and that it is impractical to ascribe every added passage to a specific layer of the DtrH. We therefore make a basic distinction between the original Deuteronomy, dtr revision as well as post-dtr revision in the context of the edition of the Pentateuch.

In the second part, we apply the methodology to additional biblical texts whose attribution is uncertain, aiming to examine the similarity of their writing style to the three ground-truth corpora. These are the late Abraham material and early Jacob story in Genesis, the Ark Narrative in 1 and 2 Samuel, and the chapters in Judges, 1 and 2 Chronicles, Esther, and Proverbs. Our results and the subsequent discussions highlight the potential of our method to shed light on contested issues in biblical studies.

## Algorithmic apparatus

In this paper, we propose an algorithmic framework for the study of (Hebrew-written) biblical texts consisting of the following stages:

1. **Lemma extraction**. Transform the words of a given biblical text to their Hebrew lemma form, utilizing predefined labeling of the Bible.

2. **Dictionary formation**. Construct a dictionary that consists of thousands of the most frequent lemmas or n-grams (i.e., combinations of several consecutive lemmas). Henceforth we commonly refer to lemmas and their n-grams as features.

3. **Text descriptor construction**. Describe each text by a frequency table, indicating the number of occurrences of each feature within it.

4. **Document-corpus discrepancy calculation**. For a pair of texts (a document and a corpus), compute a relative discrepancy score, called HC-discrepancy [22], based on the frequency table of each text. A low score indicates that it is likely that both texts were written by the same author, while high values indicate authorship discrepancy.

5. **Attribution likelihood assessment**. Given a reference corpus and a new text, propose a model for the distribution of HC values, then test the null hypothesis $H_0$, that *the given text and the corpus were written by the same author*. The calculated p-value ($p$) indicates the significance of the association of the current text to the given corpus with respect to the distribution of the HC scores of the corpus. If $p \leq 0.05$, we reject $H_0$ and accept the competing hypothesis of two different authors; otherwise, we remain undecided as we cannot rule the possibility that the new text was written by the corpus's author. The two texts may still be different in terms of authorship but our evidence is not sufficient to determine that hence we remain undecided.

6. **Authorship verification and attribution**. Given a document of unknown authorship and several candidate corpora of different authorship, calculate the p-value as described in step 5 for each corpus. Among the corpora that are not rejected with a significant 0.05, attribute the document to the most likely author. If no such corpus exists, declare that the document cannot be associated with any of the candidates.

7. **Interpreting authorship distinction**. Analyze the method's decision for two different authors by identifying sets of features that drive the highest HC-discrepancy.

Steps 1-3 above are associated with pre-processing while Steps 4-7 with inference. The result of Steps 5-6 is a table containing the p-values summarizing the evidence for the verification (proximity/discrepancy) of each text to one of the corpora. The novelty of our approach is the attribution of texts pertaining to the Hebrew Bible in an interpretable manner, that delivers accurate results even for relatively short texts. The power of our methodology lies in the following three stages (a) Assessing the likelihood of attributing a new text to a given corpus. (b) Finding the corpus for a new text that is more likely to be written by the same scribe. (c) Providing the reasoning for the text attribution (a list of features). To the best of our knowledge, there is no published automatic analysis of biblical texts that provides an interpretable attribution of biblical texts on the scale and in a rigorous manner as reported herein.

Noteworthy is the life cycle of the term *Higher Criticism*, originally used in a branch of philology that investigates the origin of a text, especially the text of the Bible [31]. In our case, Higher Criticism is a term coined by the statistician John Tukey to describe a tool for meta-analysis of multiple hypothesis testing situations [38]. In this paper we apply Tukey's Higher Criticism to analyze frequency tables; we do not consider the philological meaning of Higher Criticism.

## 2. Materials and methods

In this section, we describe the data and the detailed implementation of the algorithmic framework. In the first part of our study, we used 50 biblical chapters pertaining to three corpora based on biblical exegesis considerations: Deuteronomy (D), Deuteronomistic History (DtrH), and Priestly texts (P). The second part of our research focused on estimating the proximity of nine groups of chapters to the three ground-truth corpora. A full list of these texts is given in the Dataset section below.

### 2.1 Dataset

In this study, we examined 50 chapters of the Hebrew Bible using the procedure outlined above. Each text from our dataset was accompanied by a label that associates it with one of the three corpora (D, DtrH, and P), classified according to biblical exegesis considerations. The ground-truth chapters were selected based on the following considerations. 1) they contain a sufficient number of words for the statistical examination; 2) they feature minimal stratigraphy (layers); 3) their classification into the three corpora is broadly agreed upon among biblical scholars. The three corpora consisted of the following chapters: (a) **Deuteronomy**: Deut 6; 12–13; 15–16; 18–19; 26; 28; (b) **Deuteronomistic History**: Deut 8–11; 27; Josh 1; 5; 6; 12; 23; Judg 2; 6; 2 Sam 7; 1 Kgs 8; 2 Kgs 17:1–21; 22–25; (c) **Priestly**: Gen 1; 17; Exod 6; 16; 25–31; 35–40; Lev 1–3; 8–9. The length of these texts appears in Table S1-S2 and Figure S1 in S1 Appendix.

Determining sufficiently long biblical texts – sparingly edited, redacted, or added to – that can be attributed to the D corpus was a challenging task. Therefore, only nine clear-cut attributed texts were selected. Still, though there is a difference in the amount of data available for the D corpus as opposed to the DtrH and P corpora, the total number of lemmas in this corpus was sufficient for the statistical analysis performed in this paper. For the widely accepted reconstruction of an original layer of Deuteronomy (D) dating back to the 7[th] century BCE, we follow Preuss (see [29], pp. 46-61). As for DtrH, only the passages that are DtrH compositions were selected (cf. [4]). Texts like the royal annals or other pre-DtrH documents that were used to write the Deuteronomist History were not selected. Finally, for the priestly texts (P), we follow Nihan's hypothesis [32] for the end of P$^g$ and avoid the later priestly texts P$^s$ and H. Even if there is a debate on the distinction between P$^g$ and P$^s$, especially on the texts of Exod 35-40 and Lev 1-3; 8-9, all scholars agree in attributing these texts to the priestly milieu P. For more details regarding the division into discrete sources see [33].

## 2.2 Feature Extraction

The application of machine learning techniques to analyze the morphology of ancient texts and scripts has seen growing interest [34-35], however, it is important to tailor the method to the medium at hand. Choosing a feature for text comparison and authorship attribution plays a key role since it should capture the unique characteristics of each text, enabling more accurate comparisons and attributions. While there exist various features that could be used (e.g. [6,11,36-38]), our analysis relies on occurrences of word n-grams which fall under the so-called "bag-of-word" approach and also are the key idea in TF-IDF [39-40]. One of the advantages of n-gram occurrences is the ability to apply statistical tools that are perfectly transparent and the relatively straightforward interpretation of the results. In this aspect, our methodology provides an advantage over previous studies that handcrafted lists of words to be considered as features and prior distributions on frequencies of individual words [6,11,14,15, 27].

Our methodological framework consists of the following several steps.

**Step 1. Lemmatization.** Extracting lemmas of each document in each corpus. Creating an automatic lemmatization is a challenging task. Usually, the developed algorithms are tailored for a specific language, with its specific suffixes and prefixes. Naturally, over the years, the English language gained a lot of attention [60-62]. A separate approach needed to be developed for Hebrew texts. In [63] the authors claim that for many morphologically-rich languages (MRLs), existing pipelines show sub-optimal performance. In recent years lemmatization algorithms were developed for Hebrew, e.g., Stanford's CoreNLP [64], or the web application created by the ONLP lab from the Open University of Israel [63], as well as analytical tools for Hebrew texts introduced by Dicta [65]. Unfortunately, from our experience, the automatic procedures produce an unsatisfactory result for our purposes.

In our case, since our research dealt with predefined texts, another solution that utilizes a lemmatized version of the Hebrew Bible was selected. The *Open Scriptures Hebrew Bible project* offered exactly the information that is required [41]). The project contains lemma and morphology data in a very convenient XML schema form (see https://github.com/openscriptures/morphhb). Each word is labeled with a unique number, referring to its lemma form (e.g. lemma number 1121 refers to the word *bani*). For the chapters and verses, our wish to compare the first step was to load both its lemmas and the morphological information for further filtering (as will be explained later).

We note that counting lemmas rather than words leads to higher counts but fewer features, as there are more words than lemmas. This situation improves the detection of deviations associated with the occurrence of a specific lemma. On the other hand, by counting lemmas we may lose a potential authorship signal: the different ways different authors use the same lemma. When texts are relatively short as in our study, the benefit of improving the signal in each feature appears to overcome the loss of information. In addition, we replace all lemmas representing proper names and gentilic nouns with a designated code. The rationale is that these types of lemmas are likely to be associated with the text's context much more than its author.

Thus, once the lemmas are extracted by the OSHB, a unique number is associated with each lemma, referring to its lemma form (e.g. lemma number 8085 refers to the word šmʿ (to hear)). This id is used to make the HC calculation easier. For example, all these words: הנשמע, ישמעו, וישמע, תשמעו, כשמעכם, שמענו (hᵃnišmaʿ, yišmʿû, wayyišmaʻ, tišmʻû, kᵉšomʻᵃkem, šᵉmāʻēnû) will have the same lemma id 8085, where the ones underlined result in two lemmas. An additional example is the words that correspond to the lemma id 7223 (translated as first): ראשנים, כראשנים, בראשנה, **והראשון** (kāriʾšōnim, riʾšōnim, wahāriʾšōn, bāriʾšōnâ). While the bolded word here results in three lemmas (ו, ה, ראשון), and the two underlined words result in two lemmas. In what follows we denote as n-grams as a combination of several consecutive lemmas. For example, ויקחו אשר צוה ו אתה, are bi-grams. We uploaded all the lemmas extracted from all the chapters to [42]. See the corresponding readme for the details about this file.

**Step 2. Dictionary Formation.** Following the lemmatization process, the morphological information is used to replace all lemmas representing proper names by a designated code (marked as <Np> in the discriminating lemma list) in order to have a joint frequency for all the proper names within a given text regardless of what these names are. We repeated this replacement procedure for lemmas representing gentilic nouns (marked as <Ng>). The rationale for these replacements: though these types of lemmas are likely to be strongly associated with the text's context, the frequency of appearance of all proper names or gentilic nouns may still be author-typical. Henceforth, we use the term "words", with the understanding that the term "word" may have a broader context in our setting (e.g., n-grams, proper names, and gentilic noun codes).

Once the texts are transformed into a list of words, we constructed a dictionary that consists of the most frequent words. Specifically, the frequency of all the words was calculated and a dictionary that contained these lemmas was created. In total, the dictionary has 1,447 unique lemmas (594 unique lemmas in D, 821 unique lemmas in DtrH, and 846 unique lemmas in P, with an overlay between the dictionaries).

**Step 3. Feature Extraction.** Counting the occurrences of each lemma in the dictionary in each text and summarizing the counts in a table (histogram). Henceforth, we refer to all entries in the table simply as "words" and to the corresponding table as a word-frequency table, with the understanding that the term "word" may have a broader context in our setting (lemma, proper names and gentilic noun codes, and lemma/code n-grams).

## 2.3 Authorship attribution via statistical analysis

**Step 4. Measuring Document-Corpus Discrepancy.** Evaluating an index of *discrepancy between two given texts.* We follow the method proposed by Kipnis and Donoho [22-24] that measures the resemblance of two word-frequency tables by means of Higher Criticism (HC) of many two-sample binomial tests. This approach leads to a discrepancy index, denoted as the *HC-discrepancy*, that is sensitive to deviations in the occurrence frequency of very few words within a potentially vast dictionary, whereas the identity of these words is unknown to us in advance [22]. The HC method, which roots laid back in the papers by Donoho and Jin in [25, 43], does not require a-priori word selection.

In what follows, we review the concept of HC-discrepancy. Later, we use the HC-discrepancy to evaluate if a new text has the same author as any of the existing corpora. See Figure 1 for the outline of the HC method and its main steps.

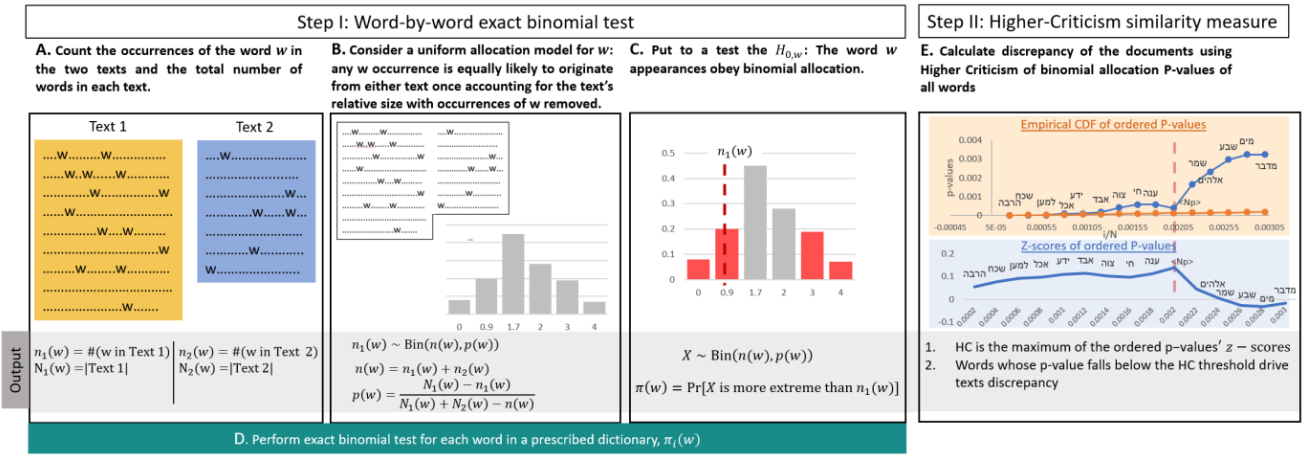

**Fig. 1 Workflow for comparing Text 1 and Text 2.** Step I: Perform exact binomial testing for each word, measuring the fit of the word's occurrences to a binomial allocation model. Step II: Conduct Higher Criticism (HC) on the per-word binomial allocation p-values and use it as an index of discrepancy between the texts. HC assesses the global significance of the p-values by comparing their z-scores to the uniform empirical process. Words associated with p-values smaller than the HC threshold are considered to provide meaningful discrimination between Text 1 and Text 2.

**Notations** Let $T = \{w_j\}_{j=1,\ldots,M}$ be a document consisting of a set of lemmas. Let us look at a corpus $\mathcal{C} = \{T_i\}_{i=1,\ldots N}$ as a set of documents, consisting of all the lemmas from all the documents along with their frequencies. We define $\mathcal{C} \setminus \{T\}$ as the corpus $\mathcal{C}$ where the document $T$ is left out, and $\mathcal{C}_T = \mathcal{C} \cup \{T\}$ as a new corpus that includes the corpus $\mathcal{C}$ and the document $T$ (the union of $\mathcal{C}$ and the corpus containing only $T$). In addition, $D \in \mathcal{C}$ denotes that the document $T$ is in corpus $\mathcal{C}$, while $D \notin \mathcal{C}$ denotes that the document $D$ is not in corpus $\mathcal{C}$. We also defined $|\mathcal{C}|$ to be the number of documents in $\mathcal{C}$.

**Definition S1. Higher-Criticism**. Given $N$ p-values $\{p_i\}_{i=1}^N$, each obtained from a different statistical hypothesis test, the Higher-Criticism measures the global significance of these p-values. It is defined as

$$\text{HC}(\{p_i\}_{i=1}^N) = \max_{1 \leq i \leq \gamma_0 N} \sqrt{N} \frac{\left(\frac{i}{N} - p_{(i)}\right)}{\sqrt{\frac{i}{N} \times \left(1 - \frac{i}{N}\right)}} \quad (1),$$

where $p_{(i)}$ denotes the $i$-th $p$-value, sorted in increasing order. Here $\gamma_0$ is a tunable parameter that typically has little effect on the large sample behavior of equation (1). We used $\gamma_0 = 0.35$. We also verified that our results are consistent across a wide range of different values of $\gamma_0$.

**Remark S1.** In words, HC is obtained by maximizing the difference between the $i$-th $p$-value and its expected value under the uniform distribution, standardized to form a z-score. The version of HC in (1) was proposed in [25], and it provided somewhat better results for the authorship context as opposed to the version of HC defined in [23], see [22] for evaluations using the two versions.

**HC-discrepancy for texts comparison**

In this section, the HC-discrepancy for comparing two texts represented by their word-frequency tables is defined and discussed. Consider a document as an ordered list of words over a prescribed vocabulary $W$. In this paper, we refer to "word" in its broader context as an $n$-gram, i.e., a combination of $n$ consecutive words in the document. For a document $T$ and $w \in W$, denote by $N(w|T)$ the number of the occurrences of the word $w$ in $T$ (Fig. 1, Step I, A). The word-frequency table associated with $T$ is the ordered pair $\{(w, N(w|T))\}_{w \in W}$.

For a given pair of documents, $T_1$ and $T_2$, we consider $W$ as the set of all words occurring at least once in any of $T_1$ and $T_2$. For each $w \in W$, we denote

$$n_w = N(w|T_1) + N(w|T_2)$$

as the total number of occurrences of the word $w$ in the two documents. Let us also define

$$q_w = \frac{\sum_{w' \in W, w' \neq w} N(w'|T_1)}{\sum_{w' \neq w \in W} n_{w'}},$$

An explanation of the motivation behind $q_w$, is provided below as part of the hypothesis $H_{0,w}$ description.

Consider the p-value $\pi(w|T_1, T_2)$ of the exact binomial test for the hypothesis

$$H_{0,w} : N(w|T_1) \sim \text{Bin}(n_w, q_w),$$

where $\text{Bin}(n, p)$ is the binomial distribution with $n$ trials and success probability $p$ of each trial (Fig. 1, Step I, B-C).

Roughly,

$$\pi(w|T_1, T_2) := \text{Prob}(|\text{Bin}(n_w, q_w) - n_w q_w|) \geq |N(w|T_1) - n_w q_w|) \quad (2)$$

($T_2$ is implicitly used in the definition of $q_w$)

Note that, due to the properties of the binomial distribution, $\pi(w|T_1, T_2)$ is commutative with respect to the order of $T_1$ with $T_2$.

The hypothesis $H_{0,w}$ corresponds to the so-called binomial allocation model of $w$ across the two documents, stating that different occurrences of $w$ are independent and each occurrence is equally-likely to originate from $T_1$ (respectively $T_2$), only accounting for the relative size of $T_1$ compared to $T_1$ minus occurrences of $w$.

The exact binomial test is applied once per each $w \in W$, result in a collection of p-values

$$\Pi(W, T_1, T_2) = \{\pi(w|T_1, T_2)\}_{w \in W}.$$

As indicated in step I, D in Fig. 1.

**Definition S2. HC-discrepancy of two documents**. Let $T_1$ and $T_2$ be a given pair of documents, and let $\pi(w|T_1, T_2)$ be the p-values obtained for each lemma that exist in the two tested documents as defined in (2). Let us define the HC-discrepancy between $T_1$ and $T_2$ with the HC score using (1) as

$$d_{\text{HC}}(T_1, T_2) \equiv \text{HC}\left(\Pi(W, T_1, T_2)\right).$$

For a numerical example see Fig. 1, Step II.

**Remark S2.** We do not rely on the correctness of the underlying binomial allocation model, as there are likely to be deviations due to dependency of some words and other regularities in the text. The HC-discrepancy is merely used here as an index of discrepancy between the two documents in the sense that large values of HC correspond to authorship discrepancy. This use of the HC-discrepancy is known to be effective in resolving authorship challenges even under violations of the binomial allocation model [22]. The correctness of the binomial allocation model is also not necessary in assessing the likelihood of attribution described in Step 5 below.

Next, the definition of the HC-discrepancy is extended in order to compare a document and a corpus of documents $\mathcal{C}$ (where $T \notin \mathcal{C}$), or between two corpora $\mathcal{C}_1$ and $\mathcal{C}_2$. In both cases, we address a corpus as a long document obtained by concatenating the content of all documents within it.

**Definition S3. HC-discrepancy of two corpora**. Let $\mathcal{C}^1 = \{T_i^1\}_{i=1,..N}$ and $\mathcal{C}^2 = \{T_i^2\}_{i=1,..N}$ be a pair of corpora. Let us define $T_1 = \cup\, T_i^1$ and $T_2 = \cup\, T_i^2$ be a new pair of documents, each document is obtained by concatenating all words in each corpus. Using definition S2, the HC-discrepancy between two corpora is defined as

$$\mathrm{d}_{\mathrm{HC}}(T_1, T_2) \equiv \mathrm{HC}\left(\Pi(W, T_1, T_2)\right).$$

Where $\pi(w|T_1, T_2)$ be the p-values obtained for each word occurring at least once in any of $\mathcal{C}^1$ or $\mathcal{C}^2$ corpora.

**Definition S4. HC-discrepancy of a document and a corpus**. Let $T$ be a document and $\mathcal{C} = \{T_i^1\}_{i=1,..N}$ be a corpus such that $D \notin \mathcal{C}$. Let us define $T_1 = \cup\, T_i^1$ the document that is obtained by concatenating all documents of $\mathcal{C}$. Using definition S2, the HC-discrepancy between $T$ and $\mathcal{C}$ is defined as

$$\mathrm{d}_{\mathrm{HC}}(T, \mathcal{C}) \equiv \mathrm{HC}\left(\Pi(W, T, T_1)\right).$$

where $\pi(w|T, T_1)$ be the p-values obtained for each word occurring at least once in any of $T$ or $T_1$.

In the paper [22] authorship study was performed on real data and demonstrated that this set typically contains more author-characteristic words than topic-related ones. Additional advantage of the HC discrepancy method is that it automatically identifies a set of words that provides the best evidence for authorship discrimination via the so-called HC thresholding mechanism [26, 43] as described in the right-hand side of Figure 1. As part of an overview of PAN authorship verification challenge in 2020 [44] the HC-discrepancy method was compared to other authorship verification methods and was found to attain competitive results. The HC method was adopted for this study due to its simplicity, interpretability, and statistical significance properties. The HC method perform well even with a small number of reference texts that are relatively short (with the shortest text consisting of around 300 words, and the median text length is 590). Additionally, it can identify discriminating signals hidden in only a few 10-30 words out of possibly thousands of words while providing statistical interpretations that are well-understood.

In **Step 5** of the method, we considered the likelihood that a new text can be associated with a given corpus. For the working hypothesis: the two texts were written by different authors, we test the null hypothesis $H_0$, that the given text and the reference corpus were written by the same author. Thus, given a document, and a corpus, we look at the extended corpus, which is the union of the two. Using the HC-discrepancies of the extended corpus, we estimate the distribution of the HC scores of the corpus (resulting in the mean and standard deviation of the HC values). Next, we propose a model: the HC discrepancies of different documents with respect to their true corpus are independent and normally distributed. We calculate a P-value $p$ that corresponds to the significance of the association of the new text to the given corpus, with respect to the estimated distribution using a t-test. If $p \leq 0.05$, we reject H₀ and accept the competing hypothesis of two different authors; otherwise, we remain undecided.

Specifically, consider a corpus $\mathcal{C} = \{T_i\}_{i=1,..N}$ of homogeneous authorship. Let us define the leave-one-out HC-discrepancy score for each document

**Definition S4. Leave-one-out HC-discrepancy score**. Let $\mathcal{C} = \{T_i\}_{i=1,..N}$ be a corpus, we define the leave-one-out HC-discrepancy score for a document $T_i \in \mathcal{C}$ and the $\mathcal{C}$ corpus as

$$x_i = \mathrm{d}_{\mathrm{HC}}(T_i, \mathcal{C} \setminus \{T_i\}) \tag{3}$$

Let $X(\mathcal{C}) = \{x_i\}$ be a set of intra-corpus leave-one-out HC-discrepancies (here $T_i$ runs over all the documents in $\mathcal{C}$). In what follows, we assume that the intra-corpus leave-one-out HC-discrepancies $X(\mathcal{C})$ for a corpus $\mathcal{C}$ of homogenous authorship are independently distributed and follows the same normal distribution. It follows from [22] that HC-discrepancies are usually less affected by topics compared to authorship, implying that the correlation between leave-one-out HC-discrepancies, if exists, is relatively small.

**Attribution Hypothesis testing.** Given a corpus $C = \{T_i\}_{i=1,..N}$ and let $T' \notin C$ be a new text, we pose the null-hypothesis $H_0$: *the new text and the reference corpus were written by the same author*. In case this hypothesis is true then the extended corpus $C_{T'} = C \cup \{T'\}$ is of homogenous authorship.

Let us first calculate the HC scores of individual documents in corpus $C$ with respect to the extended corpus $C'$

$$X(C_{T'}) = \{x_i\} = d_{HC}(T_i, C_{T'} \setminus \{T_i\}).$$

Let us calculate the HC-discrepancy score of the new document with respect to the corpus $C$ as

$$x' = d_{HC}(T', C).$$

As well as the mean and standard deviation of the HC-discrepancy scores as

$$\bar{X}(C) = \frac{1}{|C|}\sum_{i=1}^{n} x_i, \text{ and}$$

$$s^2 = \frac{1}{|C|-1}\sum_{i=1}^{n}(x_i - \bar{X}(C))^2 .$$

Then the t-statistic is defined as

$$t = \frac{x' - \bar{X}(C)}{s\sqrt{1 + \frac{1}{|C|}}},$$

where $|C|$ be the number of documents in $C$.

Under $H_0$ and provided $x'$ and $\{x_i\}$ are independent and follows the same normal distribution, $t$ follows a t-distribution with $|C| - 1$ degrees of freedom. A p-value ($P$) under $H_0$ is the probability of observing a value larger than $t$ assuming $t$-distribution using a table of values from Student's t-distribution. If this event is very unlikely (e.g., $p \leq 0.05$), $H_0$ can be rejected and we conclude that the new document is $T'$ unlikely to be written by the author of corpus $C$.

We note that it is possible to carry over (and we initially did) a similar analysis using a non-parametric test as in [22]. Nevertheless, we preferred the parametric Gaussian model because it is more robust in our context as bootstrap analysis implies.

**Step 6. Attributing Authorship**

In this subsection, we discuss the attribution process itself. Given a document $T'$ of unknown authorship and $m$ references corpora $\{C_j\}_{j=1,...m}$, each associated with a different candidate author. Noteworthy, that our analysis uses the same p-values (Step 5 of the method) for two distinct tasks: (1) hypothesis testing with respect to individual authors, i.e. authorship verification. (2) Deciding on the most likely author among the candidates, i.e. authorship attribution. In task (1), failure to reject the null hypothesis does not necessarily mean that the null hypothesis is true. It simply means that the evidence that we currently have is insufficient. It is very possible that we reject the same null given more data is available. In the context of our analysis, this means that we cannot rule out the possibility that the examined text was not written by the author against which we are testing. In task (2), we associate the text to whichever author has it most likely to be the author under the probabilistic model we derive in Step 5. It happens to be so that under the model assumptions, the most likely author also has the largest p-value.

We perform pair wise authorship attribution, and attribute the text to the most likely corpus. First, the HC-discrepancies of $T'$ is calculated with respect to each with each of the reference corpora, this process results in $x_j$. Next, the p-values ($p_j$) of associating document $T'$ with respect to the $m$ reference corpora are estimated using the t-distribution, where, $p_j \in [0,1]$. Later, the new text is associated with the corpus that has the largest p-value:

$$j^* = \text{argmax}_{j=1,\dots,m} p_j.$$

We conclude that the current text was most likely written by the author $j^*$. Nevertheless, if $p_{j^*}$ is too small, we state that the probability for the attribution is small, and we may conclude that neither of our candidate authors wrote $D'$.

**Step 7. Reasoning Attribution Procedure.** In order to identify the discriminating words, we follow the Higher Criticism Thresholding (HCT) procedure proposed in [43] The HC calculation in equation (1) considers the maximum of z-scores $z_i = (i/N - p_{(i)})/\sqrt{i/N(1 - i/N)}$. We maximize these z-scores over the range $0 < i \leq \gamma_0 N$ and denote the maximal index by $i^*$. The set of p-values selected in the HCT procedure is

$$\Delta^*(\{p_i\}_{i=1}^N) \equiv \{p_{(1)}, \dots, p_{(i^*)}\}.$$

Intuitively, HC selects a set of the smallest p-values that derive the largest deviation of the standardized empirical process represented by the $\{z_i\}$ from the uniform empirical process which arises when the p-values are uniform. In practice, we think of words or n-gram whose p-value is included in $\Delta^*(\{p_i\}_{i=1}^N)$ as features providing the best evidence against authorship similarity of the two documents. Our numerical experiments are accompanied by a list of these discriminating words and their dominance during the same authorship hypothesis test in Step 4. In addition, the methodology presented in this paper is not limited to a single word's frequencies, but also to the frequencies of n-grams (bi-gram/tri-gram).

## 2.4 Robustness assessment

We verified the robustness of our algorithmic framework using several tests. First, we evaluated the accuracy over 100 bootstrap iterations with respect to the tested data, each iteration involved: randomly sampling the entire dataset with repetitions (e.g., for each word repeated more than once we added a running index) and applying the attribution process to the sampled data where each document comprises of all words from the new sample that their index is part of that document in the original data. Using this methodology, we estimated the standard deviation of the accuracy which is 4%. Namely, the accuracy of the HC attribution method deviates by 4% from the vanilla ground-truth attribution (84%). This indicates that the method is robust with respect to small changes in the data.

Next, we raised the question of *whether the method is dependent on the ground-truth data*. We applied k-fold cross-validation, with k=4, i.e., by randomly splitting our ground-truth data (50 chapters) into four subgroups of chapters (each group consisting of 13 chapters). In each iteration, we randomly selected one group to serve as validation and used the other three groups as the reference ground-truth data (i.e., 39 chapters). Next, we attributed each document from the validation data to one of the existing corpora (D, DtrH, or P) with respect to the 39 chapters. Executing 130 random splits resulted in an accuracy mean of 85.8% with a standard deviation of 5% across repetitions. In addition, we verified that the accuracy is robust with respect to the selected feature, i.e., single term, bigram, trigram. Indeed, the accuracy remained consistent.

Finally, we evaluated the *accuracy of the HC-attribution with respect to the length of the tested texts* (in terms of the number of verses a text contains). Our analysis showed that even for relatively short chapters, 10 verses long, we reach 80% accuracy, which is excellent for such a small number of words. For additional details about these tests see Supporting Information.

# 3. Results

We analyzed 50 chapters of the Hebrew Bible using the aforementioned procedure. Each text in our dataset was assigned a label corresponding to one of the three corpora (D, DtrH, and P), categorized based on biblical exegesis criteria (see Materials and Methods for further details).

## 3.1 Likely or not: attributing ground truth data

In the first phase, we applied our approach to study the authorship of the selected 50 chapters, in a leave-one-out manner. We evaluated Steps 2-7 for each of the 50 chapters while removing a single chapter each time, and calculated the HC-discrepancy (Step 4 above) with respect to the three corpora (containing the other 49 chapters). In Fig. 2, we present a three-dimensional visualization of the chapters' embedding using the HC-discrepancy method, where each axis represents one of the three reference corpora (D, DtrH, and P). Each point corresponds to a chapter, indicating its HC-discrepancy with respect to each of the reference corpora. Table S3 in Supporting Information, shows a detailed list of all the HC values. The colors in the figure represent ground-truth attribution, where yellow was used for D, blue for DtrH, and pine green for P. Please note that the coloring serves merely as a label and does not impact the experiments in any way. Subsequently, the plots were regenerated for all three types of projection on any pair of axes. See Figure S4 in S1 Appendix for a pairwise comparison of two ground-truth corpora based on their HC values. The implication suggests an almost distinct separation between the three authors, while also indicating their proximity to each other (with only two text that were false negatively negated the attribution of biblical experts, see to the accuracy discussion below).

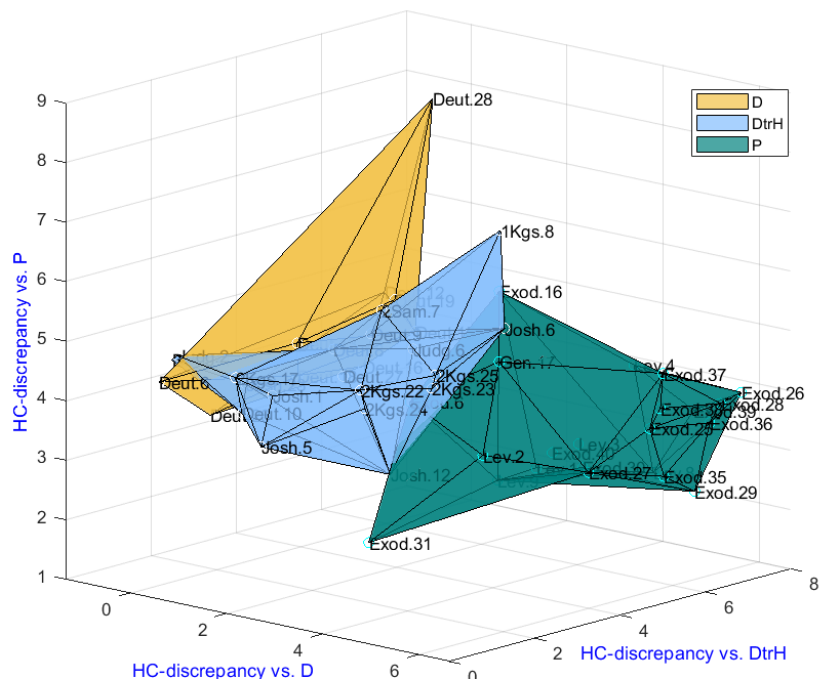

**Fig. 2. Examined biblical data displayed using the HC-discrepancy values**. Each point corresponds to a chapter, indicating its HC-discrepancy with respect to each of the corpora (D, DtrH, and P). The labels on the nodes correspond to chapter names. For validation purpose only the convex hull of the chapters was colored based on the ground-truth attribution (yellow for D, blue for DtrH, and pine green for P).

Our analysis (see Figure S4 in S1 Appendix) indicates that the examined chapters are grouped into three clear-cut clusters with a small overlap between them – each cluster echoing one of the corpora. While the first two authors (D and DtrH) are closely related, a distinct separation is evident between them and P, aligning with expert assessments. This indicates that the three corpora have distinct linguistic properties within each author cluster. Moreover, that with the very simple, yet efficient feature we were able to identify these clusters, that aligned with expert assessments.

One of the main advantages of our method is that it is fully interpretable, which means that we can trace down the reasoning of the HC embedding. In Fig. 3 we present these discriminating lemmas in a graphical manner. Each column in the graph shows a list of the 20 most discriminating lemmas for each corpus vs the union of the other two corpora as a volcano plot. The lemmas are ordered by their degree of deviation from a random allocation of occurrences across the two texts (log of the binomial allocation p-value; see SI for an explanation of the binomial allocation model). The sign of the bar indicates whether the word had high frequency in the given corpus or in those we are comparing against (positive or negative respectively). For example, discriminating lemmas for the D corpus with respect to the union of DtrH and P corpora are לא, אלהים, and <Np> (*lōʾ*, not, *ʾelōhim*, God). On the other hand, for the priestly material (the right column in Fig. 3), the lemmas מלך, אשר, אלהים, לא (*melek*, king, *lōʾ*, not, *ʾelōhim*, God, *ʾašer*, which), were highly frequent in D and DtrH but not in P, while the word זהב (*zāhāb*, gold) is a characterizing word for P. The HC calculations take into account both literary genre elements and milieu-specific theological expressions in determining the attribution of texts to one or the other corpus. The low use of relative phrases (low frequency of the word *ʾašer*, which) in P texts is a grammatical element that depends neither on literary genre nor on theology.

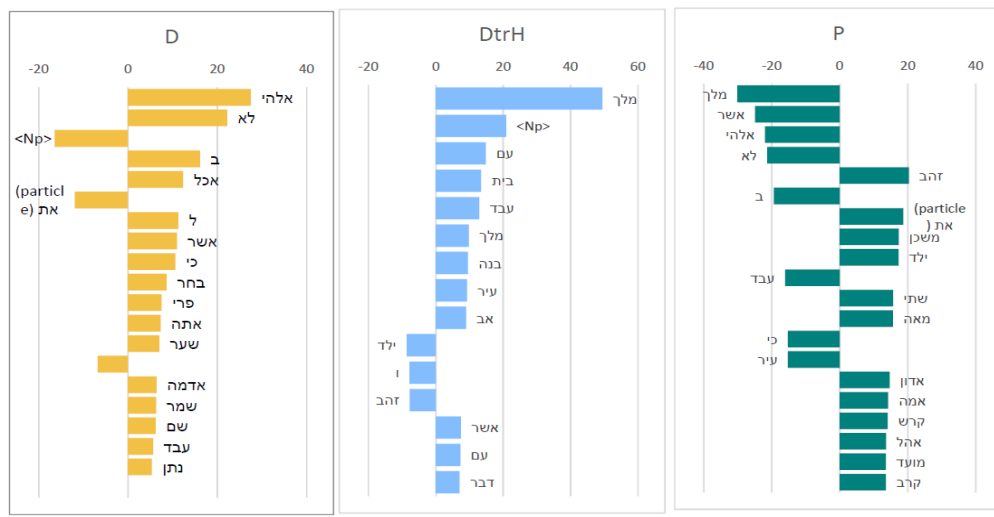

**Fig. 3. Discriminating lemmas for each corpus (D, DtrH, and P) are presented in three graphs** (left to right: D, DtrH, P). Each graph lists 20 lemmas (ordered by their importance) of each corpus vs the union of the two other corpora.

In the next step, we used the estimated three distributions of the HC scores of the ground truth corpora, and check whether a chapter was written by the same author as a specific cluster (Step 5 above). In our model, we assume that the HC discrepancies of different documents, with respect to their true corpus, are independent and normally distributed. We support our Gaussian distribution assumption by examining the goodness-of-fit test of whether sample data have skewness and kurtosis matching a normal distribution using the Jarque-Bera method. We calculated the p-value of whether the distribution of the HC masseurs of each corpus follows the normal distribution. The results are presented in Table 1, where the examined chapter titles head the columns, while the examined corpus name heads the rows, with the intersection cells providing the p-values $p$. The chapter name in the first row is colored according to the ground-truth attribution (yellow for D, blue for DtrH, and pine green for P). The table provides two types of statistically significant information: first, it indicates that for a given text, it is unlikely to belong to the corresponding corpus under the null hypothesis. Second, it indicates, with high probability, that the text can be attributed to a specific corpus. These findings address the main question in text attribution by both negating potential attributions (to one or several reference corpora) and providing positive attribution answers.

**Table 1. p-values of ground-truth chapters for the author attribution task.**

| | Deut 6 | Deut 12 | Deut 13 | Deut 15 | Deut 16 | Deut 18 | Deut 19 | Deut 26 | Deut 28 | Deut 8 | Deut 9 | Deut 10 | Deut 11 | Deut 27 |
|---|---|---|---|---|---|---|---|---|---|---|---|---|---|---|
| D | 0.97 | 0.45 | 0.64 | 0.19 | 0.38 | 0.54 | 0.47 | 0.87 | 0.23 | 0.37 | 0.16 | 0.61 | 0.90 | 0.30 |
| DtrH | 0.94 | 0.03 | 0.75 | 0.33 | 0.28 | 0.41 | 0.35 | 0.74 | 0.07 | 0.36 | 0.33 | 0.86 | 0.74 | 0.66 |
| P | 0.27 | 0.05 | 0.17 | 0.14 | 0.26 | 0.26 | 0.10 | 0.49 | 2.41E-04 | 0.10 | 0.05 | 0.41 | 0.17 | 0.23 |

| | 1 Kgs 8 | 2 Kgs 17 | 2 Kgs 22 | 2 Kgs 23 | 2 Kgs 24 | 2 Kgs 25 | 2 Sam 7 | Josh 1 | Josh 5 | Josh 6 | Josh 12 | Josh 23 | Judg 2 | Judg 6 |
|---|---|---|---|---|---|---|---|---|---|---|---|---|---|---|
| D | 0.02 | 0.53 | 0.18 | 0.14 | 0.22 | 0.08 | 0.13 | 0.34 | 0.20 | 0.58 | 0.32 | 0.05 | 0.86 | 0.18 |
| DtrH | 0.04 | 0.91 | 0.78 | 0.41 | 0.66 | 0.49 | 0.50 | 0.99 | 0.31 | 0.98 | 0.99 | 0.10 | 0.83 | 0.34 |
| P | 0.01 | 0.12 | 0.20 | 0.25 | 0.40 | 0.27 | 0.03 | 0.24 | 0.74 | 0.30 | 0.51 | 0.12 | 0.13 | 0.17 |

| | Exod 6 | Exod 16 | Exod 25 | Exod 26 | Exod 27 | Exod 28 | Exod 29 | Exod 30 | Exod 31 | Exod 35 | Exod 36 | Exod 37 | Exod 38 | Exod 39 |
|---|---|---|---|---|---|---|---|---|---|---|---|---|---|---|
| D | 0.12 | 0.12 | 2.92E-03 | 5.07E-03 | 2.56E-03 | 7.68E-04 | 1.18E-03 | 0.02 | 0.11 | 3.26E-03 | 1.77E-03 | 1.21E-03 | 1.67E-03 | 0.01 |
| DtrH | 0.25 | 0.11 | 3.40E-04 | 2.57E-05 | 3.78E-03 | 9.03E-05 | 4.96E-05 | 2.09E-03 | 0.49 | 3.55E-04 | 3.94E-05 | 3.62E-04 | 3.59E-04 | 7.35E-05 |
| P | 0.28 | 0.02 | 0.51 | 0.30 | 0.73 | 0.39 | 0.90 | 0.74 | 0.98 | 0.80 | 0.48 | 0.13 | 0.34 | 0.36 |

| | Exod 40 | Gen 17 | Lev 1 | Lev 2 | Lev 3 | Lev 4 | Lev 8 | Lev 9 |
|---|---|---|---|---|---|---|---|---|
| D | 0.04 | 0.07 | 0.01 | 0.02 | 2.90E-03 | 0.01 | 0.01 | 0.03 |
| DtrH | 0.01 | 0.11 | 0.01 | 0.07 | 4.82E-03 | 1.20E-03 | 2.05E-04 | 0.02 |
| P | 0.64 | 0.22 | 0.74 | 0.77 | 0.58 | 0.19 | 0.83 | 0.81 |

The analysis in the table suggests two outcomes. The table provide two types of information: (a) if p≤0.05, we reject $H_0$, indicating that the given text is unlikely to belong to the corresponding corpus under the null hypothesis (text colored in blue). (b) each chapter is attributed to one of the corpora (D, DtrH, P) based on the maximal p-value of the three corpora (highlighted in orange).

**Analysis 1.** We pose the null hypothesis $H_0$, *that the given text and the tested corpus were written by the same author*. if $p \leq 0.05$, we reject $H_0$, and conclude that the given text and the reference corpus are unlikely to be written by the same person under the null hypothesis (the p-value is colored in blue in Table 1 below). As can be seen, the True positive (TP) attribution rate is 0.68, 0.71, and 0.98 for the D, DtrH, and P accordingly. Thus, the assumption that the HC scores are normally distributed is justified because the spread of these scores agrees with the quantiles of the normal distribution. The False Negative (FN) occurs if a chapter is associated with a certain corpus by biblical experts but the $H_0$ was rejected for it. Except for two texts out of 50 we negated the original biblical exegesis attribution. This FN rate merely agrees with the expected FN rate of 5% under our model. The chapters that we negated their attribution erroneously are 1 Kgs 8 (that should be attributed to DtrH), and Exod 16 (that should be attributed to P). The low FN rate demonstrates the power of our methodology and opens the door to applying it to additional biblical texts, as will be described in the second part of the study reported below. Noteworthy is that we were not able to attribute 1 Kgs 8 to any of the corpora, thus we remain with 49 chapters for the attribution task.

**Analysis 2.** In the next step, we turn to the attribution task itself (Step 6 above). For each chapter, three values are presented (in the corresponding column), each indicating the likelihood of attribution to the corresponding corpus. Subsequently, for each text, we look for the corpus that is more likely to have the same authorship, i.e. the corpus with the higher probability. For each chapter, we highlighted in orange the p-value with the highest attribution likelihood. As can be seen, in 84% of the cases, the automatic attribution coincides with the biblical scholarship attribution (41 out of the 49 chapters were attributed correctly). Since the error rate is also influenced by text length (see Figure S8 in S1 Appendix), this can explain some of the misclassification. Diving in to the true positive accuracy for each corpus individually result in 78%, 72%, and 95% for D, DtrH, and P respectively (with 7 out of 9, 13 out of 18, and 21 out of the 22 chapters attributed correctly). For detailed confusion matric see Tables S4-S6 in the supporting information. This indicates that it is easier to distinguish P texts from D and DtrH texts than to distinguish D texts from DtrH texts, affirming our conclusions above regarding the word usage patterns existing in the P corpus. Noteworthy that 7 out of the 8-misclassification occurred between D and DtrH. While we select the most likely attribution, it's important to recognize that some misclassifications can arise from small differences in likelihood. For example, Josh 23 was misclassified as P and not DtrH with 0.12 vs 0.10 respectively, and Judg 2 was misclassified as D instead of DtrH (with likelihood of 0.86 vs 0.83).

In order to better understand what are the reasoning for the success attribution we need to dive to the words behind each chapter. Specifically, which words indicated that Exod 25-31 and 35-39 and corpus D (nor DtrH) were not written by the same author. For example, the first distinguishing words for Exod 25 from D (and DtrH) are זהב, טבעת, ארון, אמה, קנה and from DtrH יריעה, קרש, משכן, ה, אחד. While for Exod 26 the distinguishing words from D are קרש, יריעה, משכן, אדון, <Np>. Full interpretability details are available online in the word tables in the link [42] (along with the description in the readme file). The words that distinguish P from D and DtrH in Exodus 25-31 and 35-39 are all related to the tent of meeting, whether they be objects of worship such as the ark, parts of the tent such as curtains, boards, rings, materials such as gold, or elements of measurement such as cubits. All these terms appear several times in these chapters and elsewhere in the P texts, which are particularly concerned with objects of worship, clergy and the rites associated with them.

Herein we briefly discuss the misclassification of the 8 chapters out of the 49. These also appear in Fig. 2 as the overlap (and also in the plots of the projections on 2D in Figure S4 in S1 Appendix). The eight chapters for which the two do not agree are as follows: Deut 13, and 15 were automatically attributed to DtrH instead of D. These texts may have undergone dtr revision. (b) Deut 8, 11, and Judg 2 were attributed to D instead of DtrH. The differences between D and DtrH are in fact not so important, and both corpora share a significant vocabulary. (c) Josh 5, and 23 were attributed to P instead of DtrH. Josh 5 indeed contains passages that are inspired by priestly terms and language. For Josh 23 this attribution is mysterious since it is by all specialists considered to be "dtr." (d) Exod 16 was attributed to D instead of P. This can be explained by the fact that the P text in Exod 16 was revised later in a Deuteronomistic language.

Here are a few comments regarding the words identified: (a) If a word appears in only one text and not in the other, it can still serve as an indicator for differentiation. (b) The sign indicates whether the word was a positive indicator for separation due to text 1 or a negative indicator due text 2. (c) We list all the indicative words that could lead to the rejection of the same-author hypothesis. However, sometimes our evidence is not sufficient to conclusively determine two distinct authors (for p>0.05). For example, the words listed in the table below could suggest a rejection of the Deut 13 and DtrH same-author hypothesis, but they were not strong indicators, since the p-values for Deut 13 across the three corpora are 0.64, 0.75, 0.17 greater than the threshold. Following step 6, the maximum likelihood decision is that the chapter is associated with the DtrH.

Distinguishing between D and DtrH texts is difficult both for scholars and for the algorithm, since the two corpora are not independent. Indeed, DtrH is built on D and borrows certain expressions from it. This can explain the misclassification of Deut 13, Deut 15, Deut 8, Deut 11, Judg 2.

Misattributions between DtrH and P can be explained by the use of specific terms usually found in the other corpus. For example, the presence of the root מול (mwl, to circumcise) in Josh 5 was probably associated with Abrahamic circumcision in Gn 17 (P). In Josh 23, the repeated use of words like טוב (ṭôb, good) or אלהים (ᵉlōhim, God) may have brought to mind the P origin stories, which exclusively use this divine appellation, and Gen 1, which repeatedly declares creation to be "good". Less importantly, the word אלה (ʾēllê, these) is used to open lists, and P texts also often use this term to open a narrative or a list (12 occurrences in selected P chapters).

## 3.2 Who wrote it: shedding light on the attribution of other biblical texts

The primary analysis indicates a significant difference in word usage among the three ground-truth corpora and demonstrates high accuracy levels in the classification task. Thus, paving the way to put to a test questions regarding other biblical texts that their attribution is under dispute among biblical experts. In this section, we apply our automatic algorithmic framework to test nine groups of chapters vis a vis the three ground-truth corpora. Noteworthy, that some of these texts are with uncertain attribution, and therefore studying their attribution through the prism of our method is appealing. Here too we selected texts that contain enough statistical information, i.e., they are long enough, with minimal evidence for phases of editing. Subsequently, while our aim was to select texts that underwent minimal editing, achieving this proved challenging. Consequently, this is the primary reason for the disputed attribution of these texts by the biblical community.

The texts to be subjected to the attribution test are:

1. **Deut 4**: often considered to be one of the latest texts in Deuteronomy, integrating priestly language (see [28], p. 532-38).
2. **Lev 26:** which is the conclusion of the Holiness Code, and is considered to be a post-Dtr and post-P chapter (see [31], p. 535-45).
3. **The Ark narrative within the DtrH:** The question here is whether the two parts of the narrative originated from the same hand [45]: (a) **Ark 1:** 1 Sam 4–6; (b) **Ark 2:** 2 Sam 6.
4. **Chronicles: Chr1:** 1 Chr 12:8–15, 23–40; 22:1–29:21, and **Chr2:** 2 Chr 19:1–20:30; 29:3–31:21; 33:11–17; 34:3–7; 36:22–23. The texts of Chronicles without a parallel in Samuel-Kings are probably late compositions (4th century BCE or later) which should be distinguished from older Dtr texts (see [46] 1993, p. 23-28 or in [47]).
5. **Late Abraham material:** Gen 14–15; 20; 22; 24. Given that the different narratives of the patriarchs in Genesis were written by several redactional milieux, we wish to examine whether we can distinguish between the narratives of similar type with the same characters but different redactors. The late Abraham material seems to date to exilic or post-exilic times [48].
6. **The Gibeah story:** Judg 19–21. This story is considered one of the latest in the Book of Judges [49-50]. Some scholars, however, associate this passage with a Deuteronomistic milieu [51].
7. **The early Jacob story:** Gen 25:7, 24–26; 28:10–22; 29:15–30; 30:25–42; 31:1–22, 46–47. In Genesis, the story of Jacob is said by most scholars to contain an early (8th century BCE) storyline from the kingdom of Israel [52]. If so, this story should be distinguished from other narrative texts.
8. **Esther:** The Book of Esther. It is a late narrative text whose main plot is written by a single milieu, probably during the Hellenistic period [53]. The received text has undergone some modifications that can be observed by comparing the MT, the LXX, and the Alpha Text.
9. **Proverbs wisdom literature:** Prov 10–31. Proverbs is a literary genre different from all other texts and serves as a test case for the algorithm.

After extracting the lemmas and building a dictionary, we calculated the HC-discrepancy between each text and each of our three ground-truth corpora, and accordingly, the association likelihoods. Table 2 summarizes the resulting p-values associated with each examined text and ground-truth corpus. As before, the examined test cases head the columns of the table, while the three ground-truth corpora head the rows, with the intersecting cells providing the p-value. When $p \leq 0.05$, i.e., the tested text is unlikely to be affiliated with the given corpus at a significance level of 0.05 under the null hypothesis, the digits are colored in blue.

**Table 2. Likelihood values of the authorship attribution of the additional texts**.

|        | Deut 4 | Lev 26 | Ark 1 | Ark 2 | Chr       | Late Abraham | Gibeah   | Early Jacob | Esther   | Prov      |
|--------|--------|--------|-------|-------|-----------|--------------|----------|-------------|----------|-----------|
| Length | 1290   | 1004   | 1512  | 569   | 2081      | 3420         | 2605     | 1570        | 3066     | 7063      |
| D      | 0.29   | **0.011** | **0.038** | 0.57 | **6.400E-04** | **0.036** | **0.019** | 0.13     | 0.05     | **4.524E-06** |
| DtrH   | 0.61   | **0.020** | **0.044** | 0.84 | **5.197E-06** | **0.010** | **0.006** | 0.10     | **0.06** | **3.578E-07** |
| P      | **0.007** | **0.024** | **0.018** | 0.46 | **4.957E-08** | **2.084E-05** | **5.183E-05** | **1.664E-03** | **1.443E-03** | **1.961E-11** |

A p≤0.05, colored in blue, indicates that the text is not likely to be attributed to the ground-truth corpus under the null hypothesis. The maximal attribution probability for each text is marked in orange.

Based on the findings in Table 2, we see that: (A) *Deut 4* and the early *Jacob story* are unlikely to be associated with *P*. This adheres to the conventional wisdom in biblical studies. The early Jacob story is far earlier than P and different in style. Regarding Deut 4, the reception of priestly motives or the intertextuality created by the recourse to similar expressions with P texts does not mislead the HC calculations which consider these allusions as secondary. The text of Deut 4 seems to be composed mainly by taking up Deuteronomistic elements, hence greater proximity with the DtrH and D texts. (B) *Esther* cannot be associated with *P*. (C) *Lev 26, Ark 1, Chr, the late Abraham material*, the *Gibeah story*, and *Proverbs* are unlikely to be associated *with either of the three ground-truth corpora*. Here too the results adhere to the conventional wisdom in biblical scholarship (for example Lev 26 for more details see [54-56]): These texts are late in date and/or different in genre.

On the positive side, we observe the following possible associations: (A) *Deut 4* is more likely to be associated with DtrH or D. (B) Although we were not able to negate the closeness between *Esther* and DtrH or D, the probability for such an association is very low (6% and 5%). Therefore, Esther is unlikely to be associated with either of the three corpora. (C) For *Early Jacob* we were not able to negate an association with D; this attribution has a 13% probability. (D) Although *Ark 1* and *Ark 2* deal with the same theme, and are considered by some to be part of one text [57-58] (for a history of research [59]), Ark1 cannot be associated with either of the three corpora, while Ark2 is close to DtrH. This adheres to the interpretation of 1 as an early, pre-DtrH northern text, and Ark 2 as part of the DtrH composition [45]. The results for the texts added and compared to the three ground-truth corpora are summarized in Table S9 in S1 Appendix. See Supporting Information for two-dimensional plots of the additional materials overlaid on the ground-truth corpora for an illustration or the numerical data (Figure S5 in S1 Appendix), as well as for the list of the discriminating words (Figure S7 in S1 Appendix).

## 4. Discussion and conclusions

The aim of this study was to tackle the question of authorship attribution in biblical texts. Our primary inquiry focused on determining whether we could discern the authorship of a chapter and calculate the probability of such attribution. Usually, the attribution task is performed manually by biblical scholars. Biblical materials pose severe difficulties for computational analysis, first and foremost due to the multilayered nature (editing and additions over centuries) and different genres (e.g., "historical" narrative, legal, wisdom) of the texts. This paper demonstrates that a computer-assisted method can find patterns and can offer new insight into authorship studies of biblical materials. The novelty of our approach is the attribution of texts from the Hebrew Bible in an interpretable manner, that delivers accurate results even for relatively short texts. The computational framework solves the authorship challenge by assessing the likelihood of attributing a new text to a given corpus; finding the corpus for a new text that is more likely to be written by the same scribe, and providing the reasoning for the text attribution. To the best of our knowledge, there is no published analysis of biblical texts that provides an interpretable attribution on the scale performed in this paper.

In our study, we analyzed the authorship of 50 chapters attributed by biblical experts to three ground-truth corpora (D, DtrH, and P). Our research sheds light on the authorship attribution of these chapters and additional disputed ones. Firstly, we demonstrated that the three corpora can be differentiated based on word frequency usage. Using the HC-discrepancy method to embed the biblical chapters into a three-dimensional space, we identified three almost distinct clusters, each corresponding to authors from D, DtrH, or P. Thus, pointing out that each author has a unique linguistic fingerprint. Notably, we also observed that the classes D and DtrH are close to another, while P is rather distinct. This indicates that the word usage patterns in the first two corpora are more similar, while those in P are different.

Secondly, after modeling the distributions of the three ground truth corpora, we tackled the question whether the given text was written by the same author as the tested corpus. This resulted in only 4% false rejection of the ground truth author. This indicates that our constructed model is a good estimate to describes the ground truth corpora.

Thirdly, we utilized the models to assess the likelihood that a text can attributed to one of the reference corpora. Using the highest likelihood value for the attribution task resulted in an 84% success rate. Specifically, the success rates for attributions to D, DtrH, and P, were 78%, 72%, and 95% respectively. These attribution results were also confirmed through examination of multiple n-grams (i.e., bigrams, trigrams), see Table S10 in S1 Appendix.

Fourth, the statistically significant results pave the way to applying the method to additional biblical texts, some with disputed attribution. In one case, we examined the Ark Narrative, and our study sheds light on a disputed issue, while Lev 26 has affinities with the priestly texts and Deut 28 (belonging to D) from which the redactors borrow expressions. The use of lexemes in the HC calculations could have been misleading in associating this chapter with either P or D. Nevertheless, the result shows that taken as a whole, the chapter is different from both corpora as argued by biblical scholars.

All our results, in the four steps mentioned above, are accompanied by discriminating lemmas that narrate the story of attribution or explain why the attribution occurred or why it was rejected. The interpretability of our method, through automatic identification of word usage patterns, is the cornerstone of our collaboration with biblical experts. This enables them to comprehend the rationale behind the algorithm.

However, transparency is essential for at least three reasons, as these factors may impact the results, although not substantially alter them. These variables include a) the length of the attributed text, b) corpus richness, and c) the similarity between corpora. Naturally, the statistical richness of the chapter, determined by its number of verses and words, significantly influences the attribution task. Our analysis revealed that for texts with fewer than 10 verses, the average likelihood of correct attribution is very low, while for texts with 10-30 verses, we typically observe, on average, around an 80% chance of correct attribution. In our analysis reported herein, the shortest chapter was Josh. 5 (with 15 verses), while the longest was Deu. 28 with 69 verses. The median chapter length was 28, which, on average, should result in an 84% accuracy rate (see Figure S8 in S1 Appendix for a detailed analysis). This underscores the remarkable nature of the results presented in this paper, given the average potential attribution accuracy that can be achievable. Indeed, we observe that the shortest chapters are more prone to incorrect attribution (e.g., Deut 13, Deut 15 - with lengths of 22-23, Josh 5 and 23 - with 15-16 verses, and Judg 2 with 23 verses, as shown in Table 1).

Another factor to consider is the richness of the reference corpus. Our analysis revealed that attributing to the richest corpus. The reference corpora differ in the amount of available data to learn the word pattern (P with 22 long chapters with median number of verses 35, vs 8 mostly very short texts with 22 verses in D with median length of 22 verses, and 20 chapter in DtrH with 25 verses). It is reasonable to assume that the reachness of the P corpus provide a more robust characterization of the writer. Additionally, there is the third factor to consider: corpus D and the DtrH corpus were relatively close in the three-dimensional space (Fig, making classification between them more challenging. For instance, 4 out of the 28 chapters (14%) were misclassified among these two corpora.

This article we demonstrate that investigating biblical texts through word/bi-gram/tri-gram frequency statistical analysis is a promising research track. Our methodology makes textual criticism of biblical texts more accessible to non-specialists and provides interpretable automatic attribution. This research opens the door to tackling additional questions in biblical studies, leading to a better understanding of the formation of the Bible.

## Acknowledgments

We wish to thank David Donoho for the fruitful discussions and his valuable comments. Alon Kipnis was supported in part by funding from the Koret Foundation and the NSF under Grant No. DMS-1816114. Shira Faigenbaum-Golovin is grateful to the Eric and Wendy Schmidt Fund for Strategic Innovation, and the Zuckerman-CHE STEM Program, the Simons Foundation under Grant Math+X 400837, and Duke University for supporting her research. The kind assistance of Mira Kipnis is greatly appreciated.